\documentclass{article}
\pdfoutput=1

\usepackage[preprint, nonatbib]{neurips_2021}

\usepackage[dvipsnames]{xcolor}
\usepackage[utf8]{inputenc} 
\usepackage[T1]{fontenc}    
\usepackage{hyperref}       
\usepackage{url}            
\usepackage{booktabs}       
\usepackage{amsfonts}       
\usepackage{nicefrac}       
\usepackage{microtype}      
\usepackage{xcolor}         

\usepackage{amsmath}
\usepackage[subpreambles=true]{standalone}     
\usepackage{tikz}           
\usetikzlibrary{calc, positioning}
\usepackage{pgfplots}
\usepgfplotslibrary{groupplots}
\pgfplotsset{compat=1.15}
\usepackage{todonotes}      
\usepackage[capitalize, noabbrev]{cleveref}
\usepackage[ruled, noend]{algorithm2e}
\usepackage{bm}
\usepackage{siunitx}

\usepackage{amsmath}

\DeclareMathOperator*{\argmin}{arg\,min}

\usepackage[backend=biber, style=authoryear, maxbibnames=100, maxcitenames=1, natbib, doi=false, url=false, isbn=false]{biblatex}
\title{Instance-wise algorithm configuration with graph neural networks}

\author{%
  Romeo Valentin\\
  Dept. of Mathematics\\
  ETH Zürich, Switzerland\\
  romeo.valentin@math.ethz.ch\\
  \And
  Claudio Ferrari\\
  Dept. of Computer Science\\
  ETH Zürich, Switzerland\\
  claudio.ferrari@inf.ethz.ch\\
  \And
  Jérémy Scheurer\\
  Dept. of Computer Science\\
  ETH Zürich, Switzerland\\
  jeremys@ethz.ch\\
  \And
  Andisheh Amrollahi\\
  Dept. of Computer Science\\
  ETH Zürich, Switzerland\\
  amrollaa@ethz.ch\\
  \And
  Chris Wendler\\
  Dept. of Computer Science\\
  ETH Zürich, Switzerland\\
  chris.wendler@inf.ethz.ch\\
  \And
  Max B. Paulus\\
  Dept. of Computer Science\\
  ETH Zürich, Switzerland\\
  max.paulus@inf.ethz.ch\\
}

\begin{document}

\maketitle

\begin{abstract}
We present our submission for the \emph{configuration task} of the Machine Learning for Combinatorial Optimization (ML4CO) NeurIPS 2021 competition. The configuration task is to predict a good configuration of the open-source solver SCIP to solve a mixed integer linear program (MILP) efficiently. We pose this task as a supervised learning problem: First, we compile a large dataset of the solver performance for various configurations and all provided MILP instances. Second, we use this data to train a graph neural network that learns to predict a good configuration for a specific instance. The submission was tested on the three problem benchmarks of the competition and improved solver performance over the default by 12\% and 35\% and 8\% across the hidden test instances. We ranked 3rd out of 15 on the global leaderboard and won the student leaderboard. We make our code publicly available at \url{https://github.com/RomeoV/ml4co-competition} .

%
%
%
%
%
%
%
%
%
%
\end{abstract}

\section{Introduction}%
The Machine Learning for Combinatorial Optimization (ML4CO) NeurIPS 2021 competition\footnote{\url{https://www.ecole.ai/2021/ml4co-competition/}} aimed at improving state-of-the-art combinatorial optimization solvers by replacing key heuristic components with machine learning models. In particular, the competition posed three tasks to improve the performance of the open-source solver SCIP \citep{achterbergSCIPSolvingConstraint2009} for solving mixed integer linear programs (MILPs) efficiently. A MILP is a linear program $\max_x w^T x$ s.t. $A x \leq b$ and $x \geq 0$ where some entries of $x$ are integer valued. Our submission addressed the \emph{configuration task} of the competition, which is an example of instance-wise algorithm configuration \citep{10.5555/1860967.1861114}. For a given MILP instance $i \in \mathcal{I}$, the task is to predict a good configuration $c \in \mathcal{C}$ for the parameters of the solver, such that the MILP can be solved efficiently. In particular, performance of the solver on a given instance $i$ and a configuration $c$ is measured by the primal-dual integral $\gamma_{ic}$ after running the solver for 15 minutes. \looseness-1

Choosing a good configuration for the solver is challenging, because the number of potential configurations for the solver is large. At the time of the competition, the SCIP solver contained more than 2,500 parameters with binary, integer or continuous domains. The number of possible configurations $|C|$ for the solver is exponential in the number of these parameters. Moreover, many conditional dependencies exist between the solver's parameters, which makes exploring $\mathcal{C}$ difficult. Such dependencies may be obvious, for example using lock fixings for the clique heuristic (\texttt{heuristics/clique/uselockfixings}) is ineffective, if the heuristic is deactivated (\texttt{heuristics/clique/freq}$=-1$). But others could be less certain or only applicable to a specific instance $i$: For example the quality of pre-solving may affect the effectiveness of separation.

Leveraging machine learning for the configuration task is difficult for at least two reasons. First, models that directly operate on the variables and constraints of a MILP should be invariant to the order of those and able to handle MILPs of varying size. Second, the solver performance is likely a discontinuous function of the MILP. That is, small perturbations of $A, b$ and $w$ can lead to jumps in $\gamma_{ic}$. For example, a slight perturbation of a constraint's coefficients could render this constraint trivial or make the MILP infeasible, likely having a strong effect on the solving process.

We address these challenges by posing instance-wise algorithm configuration as a supervised learning problem. Firstly, we compile a large dataset of the solver performance for various configurations and all provided MILP instances. To facilitate this, we exploit {domain knowledge} and use a simple {greedy search} together with a metric to measure the quality of a configuration space that we propose. In this way, we reduce the number of configurations under consideration to 40 -- 60 effective configurations. Secondly, we use this data to train a graph neural network that learns to predict a good configuration for a specific instance. Our model uses the bi-partite graph representation of a MILP from \citet{gasseExactCombinatorialOptimization2019}. It is invariant to the order of constraints and variables and can handle MILPs of varying size. For a given instance, our model predicts the solver performance of all configurations under consideration simultaneously. It is trained with a simple MSE loss and learns from the solver performance of \emph{all} configurations under consideration. At test time, only a single forward pass for an instance $i$ is required and the configuration with the best predicted performance is chosen. 

Our submission was evaluated on the three problem benchmarks from the competition, \emph{item placement}, \emph{load balancing} and \emph{anonymous}. The first two contain 10,000 industrially-sized MILP instances for training and validation, while the latter contains only 117 instances that differ widely in size and are from diverse sources. The test sets were private at the time of submission and only released subsequently. We improved solver performance over the default configuration by 12\%, 35\% and 8\% respectively across all test instances. Our submission ranked 3rd (out of 15) on the global leaderboard and won the student leaderboard. We make the code of our submission available at \url{https://github.com/RomeoV/ml4co-competition} and hope that it will facilitate further research into instance-wise algorithm configuration with graph neural networks.

\section{Method}
\begin{figure}
  \begin{tikzpicture}
    \node[draw, fill=blue!5] {\includegraphics[width=\textwidth]{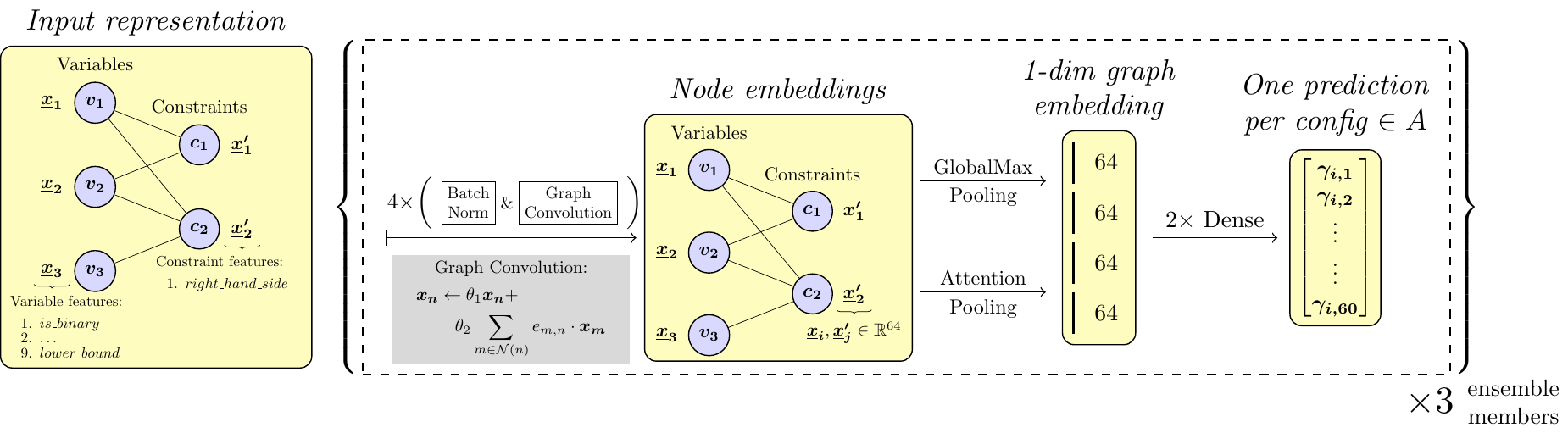}};
  \end{tikzpicture}
  \caption{Left: Bi-partite graph representation of a MILP instance $i$ from \citet{gasseExactCombinatorialOptimization2019}. Right: Our graph neural network. We follow \citet{gasseExactCombinatorialOptimization2019}, but use four convolutional layers, the graph convolution operator from \citep{morrisWeisfeilerLemanGo2019} and batch normalization \citep{ioffeBatchNormalizationAccelerating2015}.}
  \label{fig:model}
\end{figure}
\begin{figure}
  \begin{tikzpicture}
    \node[draw, fill=blue!5] {
      \includegraphics[width=\textwidth]{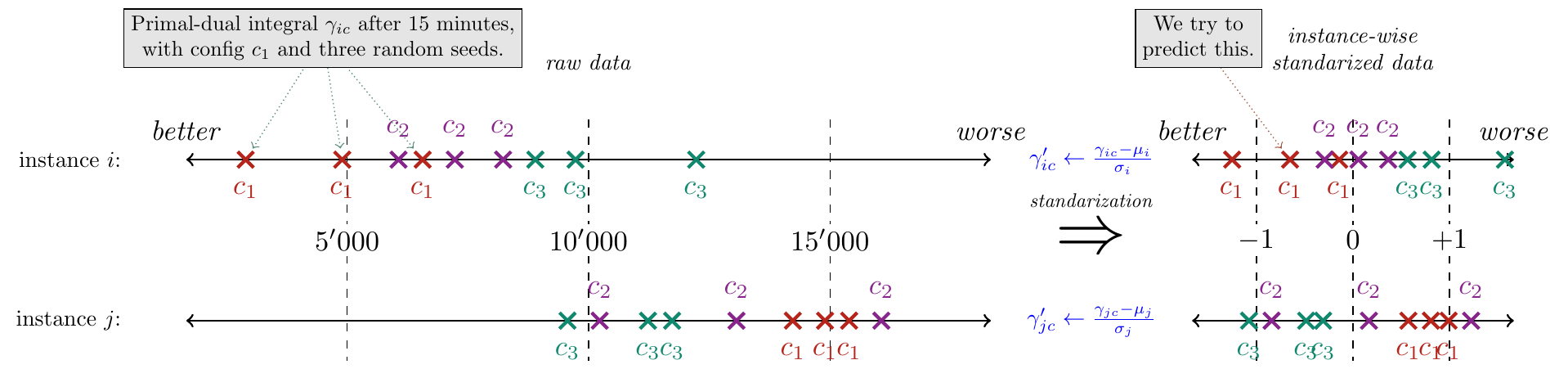}
    };
  \end{tikzpicture}
  \caption{Instance-wise target standardization: We standardize the primal-dual integral $\gamma_{ic}$ for each instance individually (here two instances $i$ and $j$). In practice, we collect multiple primal-dual integrals for each $(i, c)$ using different random seeds.}
  \label{fig:instancewise-normalization}
\end{figure}

Our method poses instance-wise algorithm configuration as a supervised learning problem. For this, we first collected a dataset of triplets $(i, c, \gamma_{ic})$ by running the solver for a fixed time limit of 15 minutes for all instances $i \in \mathcal{I}$ and all configurations $c \in \mathcal{S}$, where $\mathcal{S} \subset \mathcal{C}$ is a reduced configuration space to record the primal-dual integral $\gamma_{ic}$. Afterwards, we used this data to train a model that predicts the best configuration for a given instance. For item placement and load balancing, we used graph neural networks; for anonymous we used a simple model based on clusters. Below, we explain the key components of our approach in more detail. 

\paragraph{Domain knowledge} Running the solver for each $(i, c) \in \mathcal{I} \times \mathcal{C}$ is prohibitively expensive, because the size of $\mathcal{C}$ is too large. Therefore, we only considered the three emphasis parameters for presolving, heuristics and separation (each with 4 levels) and the built-in emphasis setting (11 levels). Each emphasis parameter sets multiple parameters of the solver consistently and simultaneously. This makes combinations of these parameters more likely to be discriminative. The Cartesian product of these four parameters results in a configuration space of size  $4^3\times 11 = 704$. By manual inspection, we observed that the Cartesian product contains duplicate configurations, whose removal produces a reduced configuration space $\mathcal{C}'$ of size $353$. For the anonymous dataset, we chose $\mathcal{S} = \mathcal{C}'$.

\paragraph{Greedy search } Given our computational budget, $\mathcal{C}'$ contained too many configurations to run the solver on all pairs $(i, c)$ when the number of instances was large. Therefore, for item placement and load balancing, we exhaustively collected data for all $c \in \mathcal{C}'$ configurations only for the first 100 training instances ($\mathcal{I}_{100}$). We then measured the \emph{quality} of a subset of configurations $S \subseteq \mathcal{C}'$ by the average solver performance that was achieved if the best available configuration in the set $S$ is selected for each instance $i \in \mathcal{I}_{100}$, 
\begin{equation}
\label{eq:goodness}
q(S) = -\frac{1}{100}\sum_{i \in \mathcal{I}_{100}} \min_{c\in S} \gamma_{ic}.
\end{equation}
We produced a sequence of subsets $S_0, S_1, S_2, \dots, S_{353}$ where $S_0 = \emptyset$, $S_i \subseteq S_{i+1}$, $|S_i| = i$ and $S_{353} = \mathcal{C}'$ by greedily adding a configuration $c^{i} \in \left(\mathcal{C}' \setminus S_{i}\right)$ to $S_{i}$ that yields the largest quality improvement, i.e., $c^{i} = \arg \max_{c \in \left(\mathcal{C}' \setminus S_i\right)} g\left(S_{i} \cup \{c\}\right)$. We observed that for item placement $q(S_{60}) \approx q(\mathcal{C}')$ and for load balancing $q(S_{40}) \approx q(\mathcal{C}')$. We thus chose $\mathcal{S} = S_{60}$ for item placement and $\mathcal{S} = S_{40}$ for load balancing to effectively reduce the number of configurations. Our procedure effectively eliminates configurations that are strictly dominated by others and ensures that there is at least one good configuration for each instance $i \in \mathcal{I}_{100}$.

\paragraph{Model} For item placement and load balancing, we designed a model that predicts the best configuration $c^{*}_i = \argmin_{c \in \mathcal{S}} \gamma_{ic}$ for a given instance $i$. Our model takes as input the bi-partite graph representation of an instance $i$ from \citet{gasseExactCombinatorialOptimization2019}. It outputs $\hat{\gamma}_i \in \mathbb{R}^{|\mathcal{S}|}$ where each entry corresponds to the performance our model predicts for a specific configuration $c \in \mathcal{S}$. At test time, a single-forward pass is performed and the configuration $\hat{c_i}$ for which our model predicts the best performance is chosen, i.e. $\hat{c}_i = \argmin_{c \in \mathcal{S}} \hat{ \gamma }_{ic}$. Our model is a graph neural network and illustrated in \cref{fig:model}. We mostly followed \citet{gasseExactCombinatorialOptimization2019}, but use four convolutional layers, the graph convolution operator from \citep{morrisWeisfeilerLemanGo2019} and batch normalization \citep{ioffeBatchNormalizationAccelerating2015} instead of layer normalization.
The processed graph is pooled (via global max and attention, separately for variable and constraint nodes) to produce a 256-dimensional latent vector from which the final prediction is made via a dense neural network with a single hidden-layer. For our submission, we produced a 3-ensemble of this model, where predictions were averaged.

\paragraph{Loss} For training the model, we used a simple but effective loss. For a given instance $i$, our model predicts the performance for all configurations simultaneously. We compute the MSE between the predicted and recorded (in dataset) performance for all configurations. This approach leverages all $\{\gamma_{ic}, i \in \mathcal{S}\}$ we collected for a given instance, while performing only a single forward-pass of the model for a given instance. We did not adopt a ranking loss, because it can make optimization harder and we observed that  often there are several good configurations for a given instance with only negligible differences, and it may be acceptable for our model to choose any of those. However, this regression task requires our model not only to predict relative performance differences between configurations, but also whether an instance is easy or hard (which is irrelevant for selecting the best configuration). Therefore, we decide to \emph{instance-wise} standardize the primal-dual integrals used as targets for training. Specifically, we compute the mean $\mu_i$ and standard deviation $\sigma_i$ over all configurations $c \in \mathcal{S}$ for a given instance $i$ to use 
$\gamma'_{ic} \leftarrow \frac{\gamma_{ic} - \mu_i}{\sigma_i}$ as our training targets (\cref{fig:instancewise-normalization}).

\paragraph{Anonymous task}
The third dataset of anonymous presents unique challenges: Firstly, the dataset is small in size; it contains only 98 and 19 instances for training and validation respectively. Secondly, the dataset is heterogeneous, it contains instances that vary substantially in size in terms of their number of constraints and variables. In contrast, the method we presented thus far relies on training a high-capacity model that can leverage a large dataset of $(i, c, \gamma_{ic})$ triplets and exploits structural commonalities in the MILP instances it is trained on. Therefore, we opted for an alternative approach that is conceptually very simple. We identified five clusters of instances across training and validation instances that share the same number of variables and constraints. We grouped all remaining instances into a sixth residual cluster. For each cluster, we determined the configuration that resulted in the best average primal-dual integral for all its cluster members based on the $(i,c,\gamma_{ic}) \in \mathcal{I} \times \mathcal{S} \times \mathbb{R}$ we collected. At test time, for a new instance, we determine its cluster membership and predict the configuration that we identified to be best for its corresponding cluster.

\section{Results}
All submissions were evaluated by the organizers of the ML4CO competition on the hidden test instances with the same hardware set-up to determine the final ranking of all participating teams. Item placement and load balancing were tested on 100 instances with a single random seed; anonymous was tested with 20 instances with five random seeds, for a total of 100 test runs. To compare our model's configurations against the default configuration of the SCIP solver, we re-evaluated our model's configurations and the default configuration on the hidden test instances after their release. We used Hewlett-Packard m710x nodes equipped with a quad-core Intel Xeon E3-1585Lv5 processor in a distributed compute cluster, for which unfortunately the residual compute load may vary during the evaluation. However, we found that our results were comparable to the performance the competition organizer's reported for our model. The competition's ranking was based on the primal-dual integral across all test instances, i.e., $\Gamma_{\hat{c}} = \sum_{i \in \mathcal{I}_{test}} \gamma_{i\hat{c}}$. This tends to give a larger weight to instances with a higher primal-dual integral. In addition, we therefore report statistics based on the distribution $\gamma_{i\hat{c}}$ and compare against the primal-dual integral of the default configuration $\gamma_{i0}$. In particular, we compute for each instance $i$, the improvement as the reduced primal-integral $\frac{\gamma_{i\hat{c}} - \gamma_{i0}}{\gamma_{i0}}$.

\begin{table*}[t]
\sisetup{detect-weight,mode=text}  
\caption{Our model improves the total primal-dual integral across all hidden test instances ($\Gamma_{c}$) by 12\%, 35\% and 8\% over the default configuration for item placement, load balancing and anonymous. It improves performance on 66, 95 and 38 out of test instances (runs) respectively.}
\label{table:results}
\begin{center}
\begin{small}
\begin{sc}
\begin{tabular}{@{}llcrcrcr@{}}
\toprule
&& \multicolumn{2}{c}{Item placement} &  \multicolumn{2}{c}{Load balancing} &  \multicolumn{2}{c}{Anonymous} \\
 \cmidrule(lr){3-4} \cmidrule(lr){5-6} \cmidrule(lr){7-8} 
Configuration && $\Gamma_{c}$ & Wins &  $\Gamma_{c}$ & Wins &  $\Gamma_{c}$ & Wins\\
\midrule
Default &&  \num{1.54e6} & 34 & \num{2.08e6} & 5 & \num{2.96e10} & \textbf{62} \\
Ours && \textbf{\num{1.36e6}}  & \textbf{66} & \textbf{\num{1.33e6}} & \textbf{95} & \textbf{\num{2.73e10}} & 38 \\
\bottomrule
\end{tabular}
\end{sc}
\end{small}
\end{center}
\end{table*}

\begin{figure}
  \centering
  \resizebox{\linewidth}{!}{
      \input{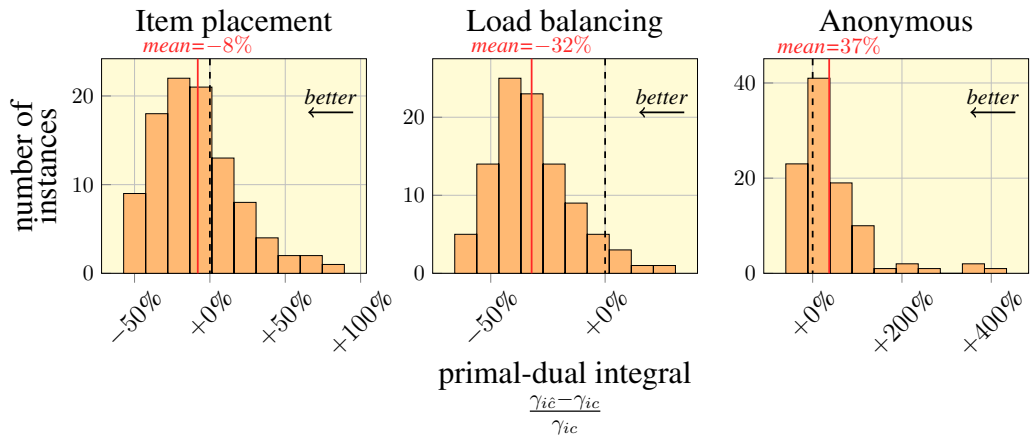}
  }
  \caption{The mean (median) improvement of our model on the test instances for item placement and load balancing is 8\% (12\%) and 32\% (34\%). But performance worsens on average (median) by 37\% (10\%) for the anonymous dataset.}%
  \label{fig:histograms}
\end{figure}

When configuring the solver with our model, the primal-dual integral across all test instances improved by $12\%$ (item placement), $35\%$ (load balancing) and $8\%$ (anonymous) respectively over the default configuration. We improved solver performance on 66 (for item placement) and 95 (for load balancing) of the 100 test instances (\cref{table:results}). For the anonymous dataset, the primal-dual integral was only reduced for 38 (out of 100) test runs. But the improvements on these instances tended to be large in magnitude, resulting in overall better performance. On the instance-level, we found that the effect of using our model's configuration instead of the default can vary widely. In the best-case, the primal-dual integral was reduced by 57\%, 66\% and 60\% respectively, while in the worst case the primal-dual integral increased by 89\%, 31\% and 434\% (\cref{fig:histograms}). The mean (median) improvement was 8\% (12\%) and 32\% (34\%) for item placement and load balancing, while performance worsened on average (median) by 37\% (10\%) for the anonymous dataset.

\section{Outlook}\label{sec:future-work}
We presented our submission for the configuration task of the Machine Learning for Combinatorial Optimization (ML4CO) NeurIPS 2021 competition. We showed that instance-wise algorithm configuration with graph neural networks is a viable approach for improving the performance of the open-source solver SCIP to solve mixed integer linear programs. The effectiveness of our approach ultimately depends on the quality of configurations under consideration. A promising avenue to improve on our work is therefore to combine our model with more sophisticated methods to explore the configuration space, possibly using online methods. We make our code available at \url{https://github.com/RomeoV/ml4co-competition} and hope that this will facilitate further research into instance-wise algorithm configuration with graph neural networks.
\printbibliography

@article{achterbergSCIPSolvingConstraint2009,
  title = {{{SCIP}}: Solving Constraint Integer Programs},
  shorttitle = {{{SCIP}}},
  author = {Achterberg, Tobias},
  date = {2009},
  journaltitle = {Mathematical Programming Computation},
  shortjournal = {Math. Prog. Comp.},
  volume = {1},
  number = {1},
  pages = {1--41},
  issn = {1867-2957},
  doi = {10.1007/s12532-008-0001-1},
  url = {https://doi.org/10.1007/s12532-008-0001-1},
  urldate = {2021-12-01},
  langid = {english},
}

@inproceedings{gasseExactCombinatorialOptimization2019,
  title = {Exact {{Combinatorial Optimization}} with {{Graph Convolutional Neural Networks}}},
  booktitle = {Advances in {{Neural Information Processing Systems}}},
  author = {Gasse, Maxime and Chetelat, Didier and Ferroni, Nicola and Charlin, Laurent and Lodi, Andrea},
  date = {2019},
  volume = {32},
  publisher = {{Curran Associates, Inc.}},
}

@inproceedings{morrisWeisfeilerLemanGo2019,
  title = {Weisfeiler and Leman Go Neural: Higher-Order Graph Neural Networks},
  shorttitle = {Weisfeiler and Leman Go Neural},
  booktitle = {Proceedings of the {{Thirty-Third AAAI Conference}} on {{Artificial Intelligence}}},
  author = {Morris, Christopher and Ritzert, Martin and Fey, Matthias and Hamilton, William L. and Lenssen, Jan Eric and Rattan, Gaurav and Grohe, Martin},
  date = {2019-01-27},
  series = {{{AAAI}}'19/{{IAAI}}'19/{{EAAI}}'19},
  pages = {4602--4609},
  publisher = {{AAAI Press}},
  location = {{Honolulu, Hawaii, USA}},
  doi = {10.1609/aaai.v33i01.33014602},
  url = {https://doi.org/10.1609/aaai.v33i01.33014602},
  urldate = {2022-01-28},
  isbn = {978-1-57735-809-1},
}

@inproceedings{ioffeBatchNormalizationAccelerating2015,
  title = {Batch {{Normalization}}: {{Accelerating Deep Network Training}} by {{Reducing Internal Covariate Shift}}},
  shorttitle = {Batch {{Normalization}}},
  booktitle = {Proceedings of the 32nd {{International Conference}} on {{Machine Learning}}},
  author = {Ioffe, Sergey and Szegedy, Christian},
  date = {2015},
  pages = {448--456},
  publisher = {{PMLR}},
  issn = {1938-7228},
  url = {https://proceedings.mlr.press/v37/ioffe15.html},
  urldate = {2022-02-10},
  eventtitle = {International {{Conference}} on {{Machine Learning}}},
  langid = {english},
}

@inproceedings{10.5555/1860967.1861114, author = {Kadioglu, Serdar and Malitsky, Yuri and Sellmann, Meinolf and Tierney, Kevin}, title = {ISAC -- Instance-Specific Algorithm Configuration}, year = {2010}, isbn = {9781607506058}, publisher = {IOS Press}, address = {NLD}, booktitle = {Proceedings of the 2010 Conference on ECAI 2010: 19th European Conference on Artificial Intelligence}, pages = {751–756}, numpages = {6} }

\end{document}